\documentclass[10pt,twocolumn,letterpaper]{article}

\usepackage[pagenumbers]{iccv} 
\usepackage{float}
\usepackage{colortbl}
\usepackage{xcolor}
\usepackage{soul}
\usepackage{arydshln}
\usepackage{pifont}
\usepackage{breqn}
\usepackage{booktabs}
\usepackage{multirow}
\usepackage{hhline}

\definecolor{iccvblue}{rgb}{0.21,0.49,0.74}
\usepackage[pagebackref,breaklinks,colorlinks,allcolors=iccvblue]{hyperref}

\def\paperID{11863}
\def\confName{ICCV}
\def\confYear{2025}

\title{Harnessing Synthetic Preference Data for Enhancing \\Temporal Understanding of Video-LLMs}

\author{Sameep Vani$^{1}$\thanks{These authors contributed equally to this work.} \quad
Shreyas Jena$^{2}$\footnotemark[1] \quad
Maitreya Patel$^{1}$ \\
Chitta Baral$^{1}$\thanks{These authors are equal advising authors.} \quad
Somak Aditya$^{2}$\footnotemark[2] \quad
Yezhou Yang$^{1}$\footnotemark[2] \\
\\
$^{1}$Arizona State University \quad $^{2}$Indian Institute of Technology, Kharagpur \\
{\tt\small \{svani, maitreya.patel, cbaral, yz.yang\}@asu.edu} \\
{\tt\small \{shreyasjena, saditya\}@cse.iitkgp.ac.in}
}

\begin{document}

\twocolumn[{%
\renewcommand\twocolumn[1][]{#1}%
\maketitle

\centering
\captionsetup{type=figure}
\includegraphics[width=\linewidth]{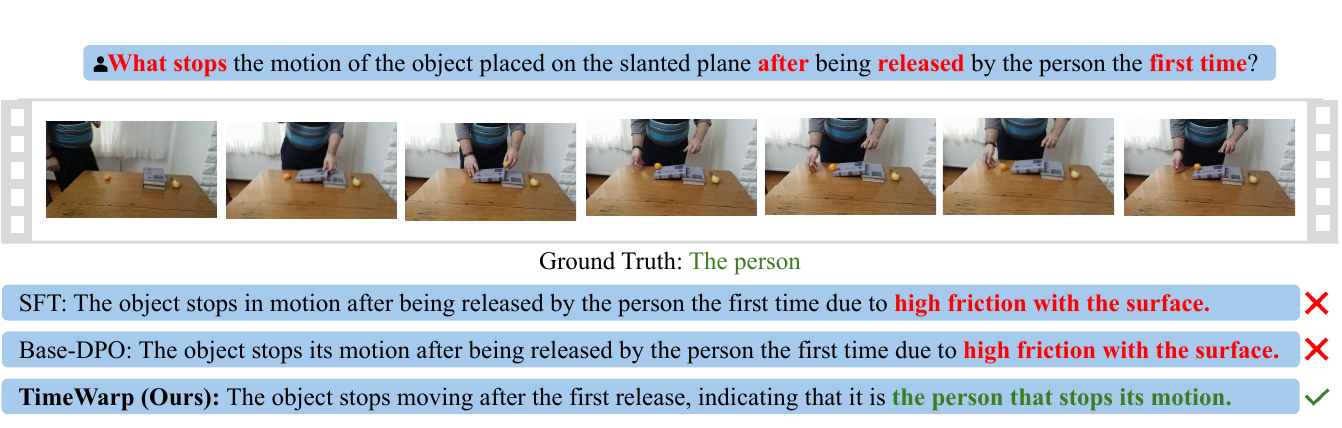}
\captionof{figure}{Existing baseline methodologies such as Supervised Fine-Tuning (SFT) and base direct preference optimization (base-DPO) that do not target temporal aspects in video exhibit a deficiency in understanding temporal dynamics, leading to challenges in capturing event order. Our method (TimeWarP) improves temporal understanding in Video-LLMs. The base model used for inference here is LLaVA-Hound.}
\label{fig:top_figure}
\vspace{8pt}
}]  

\begingroup
\renewcommand\thefootnote{\fnsymbol{footnote}} 
\footnotetext[1]{These authors contributed equally to this work.}
\footnotetext[2]{These authors contributed equally towards mentoring this work.}
\endgroup

\newcommand{\benchmark}{TimeWar}
\newcommand{\finetune}{\textbf{TimeWarp}}
\newcommand{\green}{\cellcolor[rgb]{0.702, 0.835,0.725}}
\newcommand{\gray}{\cellcolor[rgb]{0.898, 0.894, 0.894}}
\newcommand{\orange}{\cellcolor[rgb]{0.976, 0.843, 0.71}}
\newcommand{\yellow}{\cellcolor[rgb]{0.999, 0.999, 0.001}}


\begin{abstract}
While Video Large Language Models (Video-LLMs) have demonstrated remarkable performance across general video understanding benchmarks-particularly in video captioning and descriptive tasks-they consistently underperform on tasks that require fine-grained temporal understanding.
This limitation arises due to the lack of visual complexity and temporal nuance in current fine-tuning datasets, leading these models to rely heavily on language-based reasoning rather than truly understanding video dynamics. 
In this work, we propose {\finetune}, a systematic method to create a targeted synthetic temporal dataset to fine-tune the model's responses to encourage it to focus on the given input video.
We introduce a large-scale preference dataset, created using {\finetune}, that captures intricate temporal dynamics often overlooked, grounding the model's responses to visual and temporal information.
We demonstrate that when our method is applied to existing models, it significantly improves performance on temporal understanding benchmarks, highlighting the effectiveness of our proposed datasets in advancing temporal understanding in Video-LLMs, resulting in an absolute improvement in performance across seven benchmarks. Code is available at \url{https://github.com/sameepv21/timewarp}.
\end{abstract}
\section{Introduction}
\label{sec:intro}
Large language models (LLMs) have gained significant attention due to their capabilities in text understanding \cite{touvron2023llama}. Moreover, these LLMs are being increasingly used in multiple modalities, including both images \cite{liu2024improved} and videos \cite{onevision}. 
As a result, these Video-LLMs have reached a new level of state of the art performance with improvements over multiple benchmarks \cite{caba2015activitynet, bain2021frozen, xu2016msr, chen2011collecting, jang2017tgif, goyal2017something} with over 80\% accuracy on MSVD benchmark \cite{zhang2024direct}. However, these benchmarks primarily assess Video-LLMs on coarse video descriptions or on questions that can be answered using isolated frames, often achievable by image-based models alone. These models often fail under specialized benchmarks like Perception Test \cite{patraucean2024perception}, targeting the temporal aspects of the videos. 
This is because analyzing the true essence of videos requires a joint understanding of the visual and temporal dynamics of the model \cite{yun2024compositional}. Moreover, the models often overlook the complexity of the visuals, forcing them to rely on language-driven reasoning. This includes the ability of the models to understand the relationship between inter-playing events in the video.
Multiple attempts~\cite{liu2024oryx, qian2024momentor, song2024temporal, ren2024timechat, huang2024vtimellm} at combining temporal information with the video features have been proposed. 
While these approaches effectively integrate temporal information for fine-grained tasks like event detection, action localization, and event boundary matching-often relying on temporal attention-based methods to pinpoint event timestamps-they fall short of explicitly modeling the events' sequence and interconnected dynamics.

In an attempt to address these issues, this paper investigates Video-LLMs' capacity to understand temporal dynamics jointly, tackling the main challenge of high-quality data acquisition. Our method is inspired by the recent success of preference optimization using synthetic data \cite{patel2025tripletclip} on LLMs \cite{xiao2024comprehensive}, and VLMs \cite{deng2024enhancing}, which leverage preference data to improve performance on downstream tasks. However, unlike the image-only domain, the video modality introduces new challenges, as Video-LLMs must understand the input of multiple frames and the relationship/connection between multiple frames before responding to any related textual queries about the video.

Motivated by this, we develop {\finetune}, a systematic method for generating large-scale synthetic data specifically annotated for temporal aspects enhancing the model’s ability to learn nuanced time-based understanding. Most existing datasets are developed to target either overall descriptions of the videos or temporal \textit{reasoning}, failing to capture broader aspects of temporal \textit{understanding}. Our method can be applied to generate both a synthetic large-scale supervised fine-tuning dataset and a preference dataset for fine-tuning.

Utilizing a novel prompting approach using GPT-4o-Mini, we generate 15k synthetic preference data containing preferred and dispreferred responses meticulously designed to capture temporal dynamics and the relationships between events as they occur in video sequences. By incorporating techniques such as shuffled order of events and reversed entire sequence of events in the video, we enable the model to learn essential temporal dependencies within videos. In addition, we introduce a dedicated temporal benchmark, TimeWar, to assess the model’s capability to understand these dynamics, underscoring the critical importance of our fine-tuning dataset for advancing temporal comprehension in Video-LLMs.

Employing direct preference optimization (DPO) \cite{rafailov2024direct} as the basis for preference optimization and using LLaVA-Hound \cite{zhang2024direct} as our backbone, we show that the final model checkpoint after fine-tuning using data generated by {\finetune} achieves an absolute improvement across all seven benchmarks. For instance, we observe an absolute improvement of 5\% in the difficult Perception Test benchmark. An example of the different responses for this benchmark, generated from training LLaVA-Hound baseline, for various methodologies, is shown in Figure \ref{fig:top_figure}. 


\noindent We highlight our contributions as follows: 
\begin{enumerate}
    \item We present {\finetune}, a novel and systematic synthetic data generation method to improve the temporal understanding capabilities of the model extendable to both generating supervised fine-tuning (SFT) and preference data.
    \item We release a 15k temporally targeted fine-tuning preference dataset covering temporal aspects of the videos serving as a fundamental resource for video understanding, assisting in enhancing the performance of Video-LLMs.
    \item We introduce \textbf{TimeWarp - Implicit}, a novel approach for generating preference data on Video Comprehension, specifically designed to capture the temporal aspects of the input sequence indirectly.
    \item We demonstrate the effective application of DPO to improve model's temporal performance by leveraging the model feedback as reward, establishing a capability to improve performance on all seven benchmarks.
\end{enumerate}

\begin{figure*}
    \centering
    \includegraphics[width=0.8\linewidth]{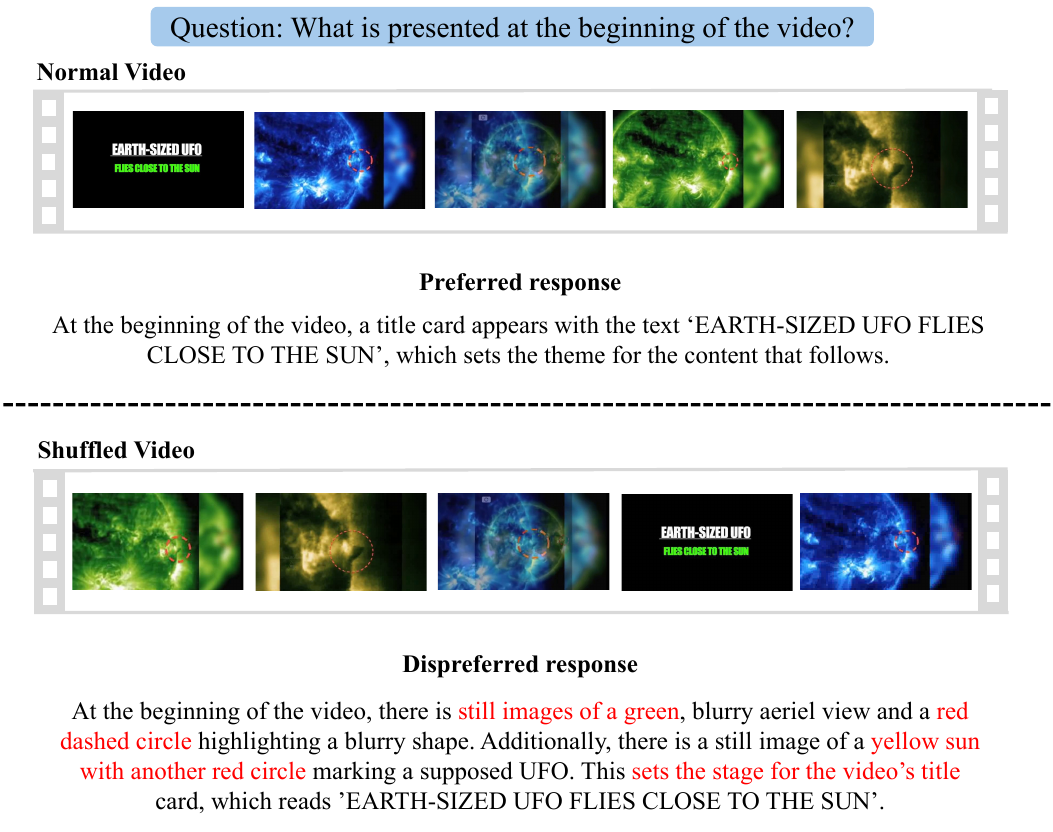}
    \caption{An example of shuffled videos and preferred and dispreferred responses generated using our proposed method.}
    \label{fig:pref_data}
\end{figure*}
\section{Related Works}
\label{sec:related_works}

\textbf{Multimodal Large Language Models}:
In recent years, the community has seen a proliferation of extensions of LLMs to comprehend images. Many attempts have been made to enforce the models to follow instructions across modalities \cite{liu2024llava} \cite{liu2024improved}. Apart from these, several works aimed at improving image understanding either in the spatial domain or in generic captions have been proposed, closing the gaps between open-source and closed-source models \cite{deitke2024molmo} \cite{wan2024contrastive} and \cite{chen2024spatialvlm}. Moreover, \cite{deng2024enhancing} introduced a self-training approach for enhancing image comprehension.
Building on the success of large-scale vision models in image understanding, there has been a rapid expansion of research focused on advancing video comprehension \cite{lin2023video, llava_next, xu2024pllava, xu2024pllava, li2024aria, onevision}. Our work leverages LLaVA-Hound \cite{zhang2024direct} as the backbone for infrastructure and baseline, enriching its temporal understanding capabilities through {\finetune}  and optimizing with Direct Preference Optimization (DPO) to capture nuanced temporal dynamics and comparing performance on multiple benchmarks including {\benchmark}. Moreover, recently, there has been a boom in models that achieve new SOTA. \cite{wang2024videoclip, qwen2.5-VL, wang2025internvideo, wang2024tarsier, damonlpsg2025videollama3, li2024videochatflash}.
In addition, recently, multiple works have proposed various methods in an attempt to deal with various temporal tasks. Momentor \cite{qian2024momentor}, Oryx \cite{liu2024oryx}, TimeChat \cite{ren2024timechat}, and VTimeLLM \cite{huang2024vtimellm} provide multiple approaches to deal with temporal dynamics of the video sequence. These approaches, including \cite{song2024temporal}, develop some sort of attention mechanism combined with image-based learning to force the model to focus on fine-grained timestamps of the video. As a result, these models are then limited to the tasks pertaining to temporal timestamps like action localization and event detection. 

\hspace{-12pt}\textbf{Video Understanding Benchmarks}:
Existing video understanding benchmarks, including Activity-Net \cite{caba2015activitynet}, MSRVTT \cite{xu2016msr}, MSVD \cite{chen2011collecting}, and WebVid \cite{bain2021frozen} focus on the generic video descriptions.
\begin{figure*}
    \centering
    \includegraphics[width=\linewidth]{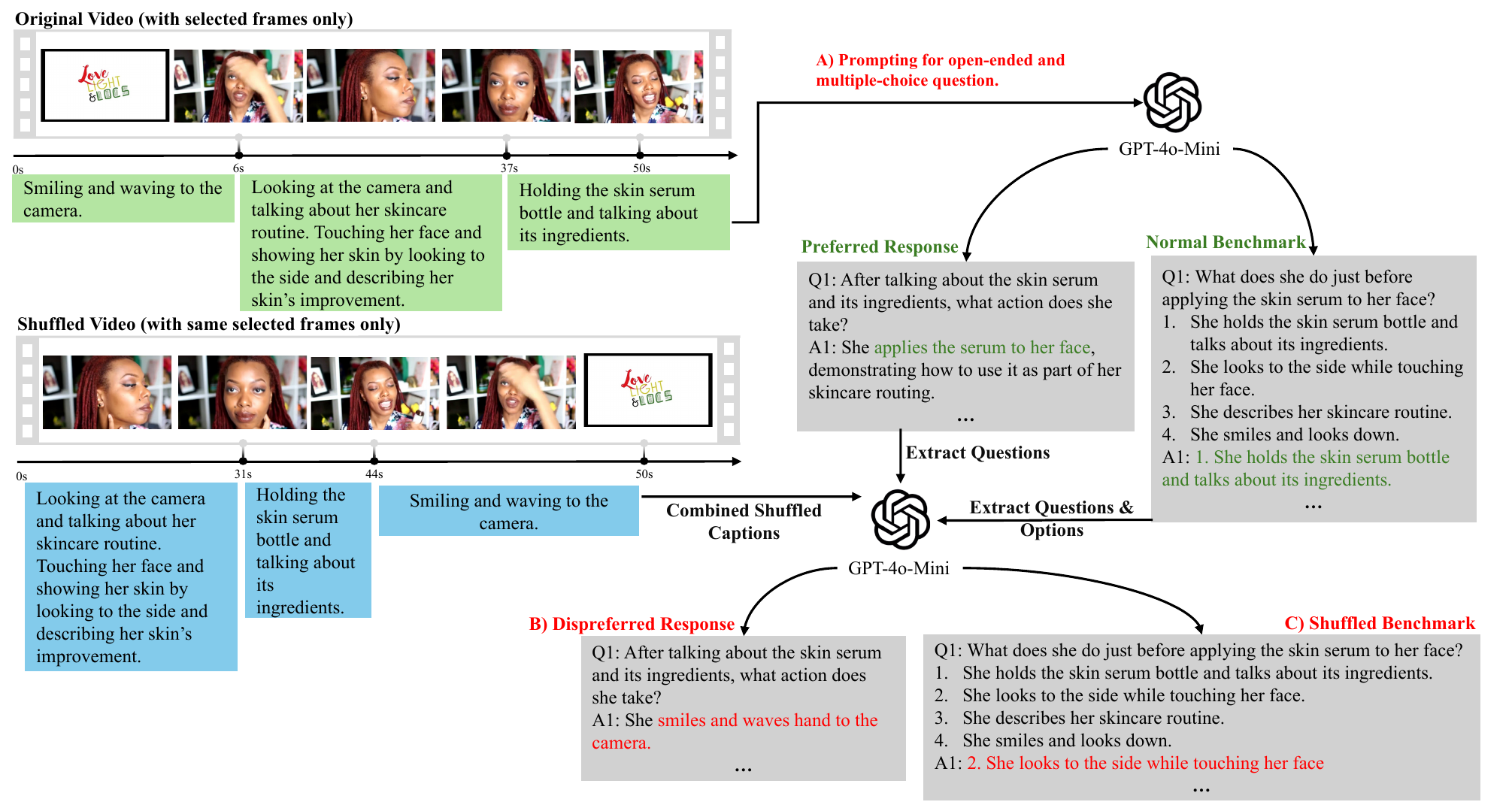}
    \caption{Workflow diagram showing the use of GPT-4o-Mini for a) generating open-ended as well as multiple choice question-answer pair; b) generating dispreferred response given the same question; c) selecting option based on the shuffled captions provided for creating a benchmark for shuffled videos.}
    \label{fig:data_pipeline}
\end{figure*}
Moreover, challenging benchmarks like LVBench \cite{wang2024lvbench}, LongVideoBench \cite{wu2024longvideobench} CinePile \cite{rawal2024cinepile}, SSV2 \cite{goyal2017something}, NextQA \cite{xiao2021next}, and ACQUIRED \cite{wu2023acquired} have been proposed that target various aspects of videos. NextQA and ACQUIRED are focused on counterfactual and causal understanding of videos. In contrast, LVBench, LongVideoBench, and CinePile are focused more on long-form or long-context understanding of the videos. SSV2 focuses on a common sense understanding of videos.

Lastly, temporal benchmarks have also been proposed recently, like Perception Test \cite{patraucean2024perception}, TGIF \cite{jang2017tgif}, and MVBench \cite{li2024mvbench} have been introduced. At some level, cinepile also models some temporal aspects. However, these benchmarks still lack explicit temporal aspects of the video, including understanding the interplay between the events. Recently, the community has seen a rise in temporal benchmarks.\cite{bagad2023test, fu2024video, bansal2024videocon, liu2024tempcompass, li2024vitatecs, zhang2024vinoground, cores2024tvbench}.

\hspace{-12pt}\textbf{Preference Optimization Techniques}:
Preliminary attempts to incorporate human feedback introduced reinforcement learning through human feedback (RLHF) \cite{christiano2017deep}, leading to the replacement of humans with AI for cost-effective scaling of the method \cite{lee2023rlaif}. These methods train a reward model to incorporate the generated feedback. To mitigate large amounts of time to first train the reward model and then train the RL model, later attempts (like DPO \cite{rafailov2024direct}, ORPO \cite{hong2024orpo} and KTO \cite{ethayarajh2024kto}) introduced methods to directly model the preference dataset as a reward model leading to much lesser overall training time. Since we are using LLaVA-Hound as our backbone, we use DPO as the optimization technique.
Myriad number of Multimodal-LLMs (MLLMs) have implemented these methods for images \cite{deng2024enhancing, zhou2024aligning, li2023silkie, wang2024mdpo}. These works have targeted multiple aspects, such as visual grounding and spatial aspects. The usage of preference optimization techniques has naturally extended to video modality. Various works \cite{zhang2024direct, llava_next, onevision} are utilizing this technique to improve the performance of various downstream tasks like grounding, action and event localization, reasoning, etc.
\section{Method}
\label{sec:method}
This section begins with briefly describing the definition of temporal understanding followed by details regarding {\finetune} for generating not only preference data but can be easily extended to generate SFT data as well. Next, we present details on TimeWarp - Implicit, our method for generating preference data indirectly. Lastly, we briefly discuss about our proposed benchmark ({\benchmark}) for evaluating models for temporal understanding. Our approach centers on constructing targeted data that emphasizes temporal dynamics within video sequences. We defer the background information on DPO \cite{rafailov2024direct} and alignment fine-tuning to the supplementary section (B).

\subsection{TimeWarp - Explicit}
\textbf{Overview and Definition}: Figure \ref{fig:data_pipeline} shows the general workflow of {\finetune} targeting temporal understanding in models.
We define temporal understanding in Video-LLMs as the model's capability to comprehend and correctly interpret events' order, duration, and relative positioning within a video sequence. This understanding requires the model to go beyond analyzing individual frames in isolation and instead integrate information across multiple frames to capture the temporal structure of the video. In this context, Temporal Understanding is evaluated through the model's ability to answer questions that explicitly target sequential relationships between events, such as determining what occurs after, before, at the beginning, or at the end of a video. Importantly, this does not involve causal reasoning, counterfactual analysis, or abstract temporal reasoning but focuses on the accurate perception and retrieval of temporal order and position within the observed video content.

\hspace{-12pt}\textbf{Base Dataset and Preprocessing}: With the above definition in mind, we utilize the FineVideo dataset \cite{Farré2024FineVideo}, which offers over 43K extensively annotated videos with an average duration of approximately five minutes, providing a diverse and temporally rich source of video data. This dataset is particularly advantageous as it offers segmented video clips with well-defined timestamps and accurate scene-level annotations. It is highly suitable for investigating and enhancing temporal understanding in video models. This allows us to isolate and analyze temporal aspects of video sequences accurately due to the structured annotations.

Since our main focus is to improve the temporal understanding of the videos rather than long-context video understanding, we pre-processed the data by trimming each video to a maximum duration of 105 seconds (1 minute and 45 seconds). This cap is strategically chosen, reflecting known limitations in handling extended temporal context in models, as identified in benchmarks like Cinepile \cite{rawal2024cinepile}. During trimming, in order to ensure the requirements of multiple scenes, we retain a minimum of two scenes per video, preserving continuity and ensuring a high-quality temporal narrative throughout the dataset.

\hspace{-12pt}\textbf{Preference Data Generation}: With these pre-processed video clips and their associated annotations, we generate composite captions that retain the correct sequence of events, maintaining the chronological flow across scenes. These sequential captions serve as inputs for GPT-4o-Mini, which we employ for QA generation. Our novel prompting method, directs GPT-4o-Mini to formulate questions focusing on temporal dynamics within the video context, ensuring each question targets a specific temporal event or transition. 
\begin{table}[!htp]\centering
\caption{Preference fine-tuning dataset statistics: Below table provides numerical statistics of dataset created using {\finetune} method.}
\label{tab:finetune_stat}
\scriptsize
\scalebox{0.95}{%
\begin{tabular}{lcccccc}\toprule
&\textbf{Total Videos} &\textbf{Avg \# of clips} &\textbf{Clip Duration} &\textbf{QA Pairs} \\
\midrule
ShareGPTVideo &300k &1 &- &17k \\
Cinepile &9396 &1 &160s &303k \\
Perception Test &11k &1 &23s &- \\
\hdashline
\addlinespace[2pt]
\orange\textbf{Ours} &\green\textbf{10k} &\green\textbf{3} &\green\textbf{21s} &\green\textbf{30k} \\
\bottomrule
\end{tabular}
}
\end{table}

\begin{table}[!htp]\centering
\caption{Preference fine-tuning dataset statistics: Below table provides qualitative statistics of dataset created using {\finetune} method. Pure Temporal denotes whether the benchmarks' target was temporal aspects or not.}
\label{tab:finetune_checkmarks}
\scriptsize
\scalebox{1}{%
\begin{tabular}{lcccc}\toprule
&\textbf{Shuffled} &\textbf{Reversed} &\textbf{Pure Temporal} \\
\midrule
ShareGPTVideo &\ding{55} &\ding{55} &\ding{55} \\
Cinepile &\ding{55} &\ding{55} &\ding{55} \\
Perception Test &\ding{55} &\ding{55} &\ding{55} \\
\hdashline
\addlinespace[2pt]
\orange\textbf{Ours} &\green\textbf{2617} &\green\textbf{2394} &\green\ding{51} \\
\bottomrule
\end{tabular}
}
\end{table}

\begin{figure*}
    \centering
    \includegraphics[width=\linewidth]{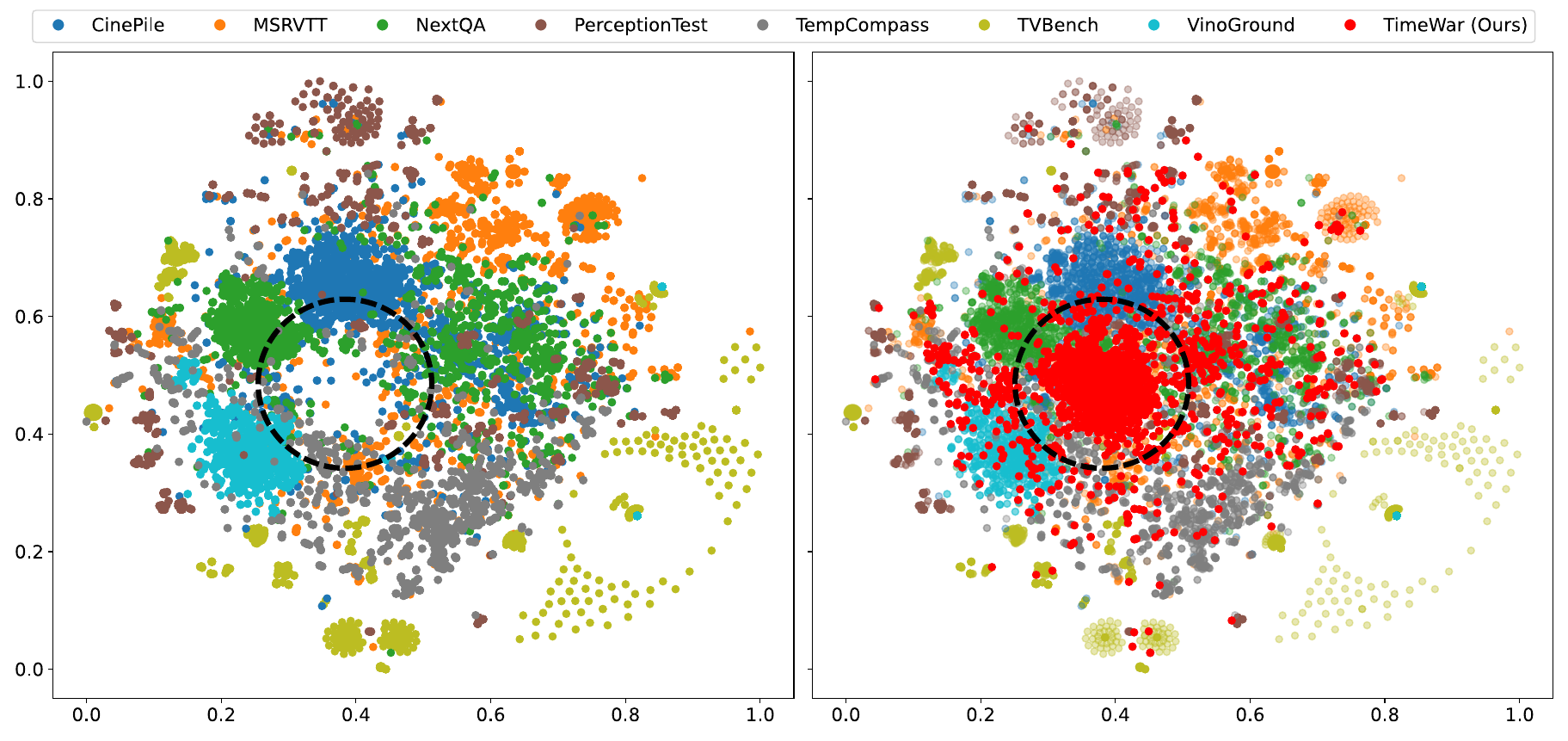}
    \caption{2D t-SNE visualization of question embeddings from {\benchmark} and seven different benchmarks, each sampling 1k random questions. Left: Shows a significant gap in current benchmarks (represented by a dotted circle). Right: Shows that {\benchmark} is able to fill the gap (represented by a dotted circle) providing the community with a much richer benchmark.}
    \label{fig:t_sne}
\end{figure*}
For generating negative captions-an essential part of Direct Preference Optimization (DPO) fine-tuning-we randomly, reorder the scenes within the video clips. In this process, scenes and annotations are shuffled while preserving the coherence of individual scenes, introducing temporal inconsistencies and challenges in the overall narrative. Unlike TVBench \cite{cores2024tvbench}, which directly shuffles the order of frames randomly without ensuring cohesion between different frames. As a result, the shuffled video feels unnatural not only for models but also for humans. Our shuffling generates intentional discrepancies that serve as hard negatives, forcing the model to distinguish between temporally consistent and inconsistent sequences and ensuring that the resulting video does not feel unnatural. As shown in Figure \ref{fig:data_pipeline}, given the modified annotations of shuffled video sequences, we prompt GPT-4o-Mini to answer questions based on these altered sequences, generating “negative” or dispreferred responses that capture temporal disfluencies. These responses form the basis for preference-based fine-tuning, enabling the model to learn and prioritize temporally coherent narratives. An example of the resulting generated preferred and dispreferred response is provided in Figure \ref{fig:pref_data}. Moreover, Table \ref{tab:finetune_stat} and \ref{tab:finetune_checkmarks} provide numerical and qualitative statistics about the generated preference data respectively.

\begin{table}[!htp]\centering
\caption{VinoGround style evaluation of models on preference data generated using {\finetune}.}
\label{tab:difficulty_analysis}
\scriptsize
\scalebox{1.2}{
\begin{tabular}{lcccc}\toprule
\textbf{Models} & \textbf{Text} & \textbf{Video} & \textbf{Group} \\

\midrule
Qwen2.5 &11.6 &7.8 &0.8 \\
InternVideo &15.4 &13.2 &1 \\
LLaVA-OneVision &17.59 &11.2 &3.2 \\
VideoChat &18.2 &11.4 &1.8 \\
LLaVA-Hound &20.2 &3.86 &2.99 \\
Video-LLaMA3 &21.4 &15.44 &1.15\\
\hdashline
\addlinespace[2pt]

\orange{\textbf{Random}} &\green{\textbf{25\%}} &\green{\textbf{25\%}} &\green{\textbf{12.5\%}} \\
\bottomrule
\end{tabular}
}
\end{table}
\hspace{-12pt}\textbf{Analyzing the difficulty of the generated data}: Let's assume we have original and shuffled videos represented as ${(V_{i}, V'_{i})}$. Let's also assume that $X$ represents the question generated by gpt-4o-mini and ${(A_{i}, A'_{i})}$ represents the preferred and dispreferred response generated by original and shuffled videos respectively. Following VinoGround \cite{zhang2024vinoground}, we measure the text-score, video-score and group-score to evaluate the baseline model as well as the SOTA models. We also provide random performance. If the preference data is truly of high-quality, then the existing models should struggle to find simultaneous correct video as well as its corresponding caption / answer and should perform close to random. Table \ref{tab:difficulty_analysis} shows that there is a huge gap between model's performance and random performance.

Note that we use {\finetune} for general cases to indicate TimeWarp-Explicit and use these interchangeably.

\subsection{TimeWarp - Implicit}
To analyze and generate high-quality temporal preference data, we propose TimeWarp - Implicit, a novel approach at generating preference data indirectly by extending Self-Training for Image Comprehension (STIC) \cite{deng2024enhancing} framework. Originally, this framework targets the lack of high quality image descriptions resulting in poor performance on multiple tasks. It employs a two-stage technique. First, it uses a synthetic data generation pipeline to generate high quality preference data. Second, it infuses the generated image descriptions with the instruction-tuning dataset to fine-tune on downstream tasks achieving impressive results in image understanding tasks. Therefore, we build upon the existing work STIC, extending it to video modality providing an additional targeted temporal preference data.

Following STIC \cite{deng2024enhancing}, we generate a self-constructed preference dataset from the base Video-LLM, which we aim to improve through preference optimization. The preferred response is the model-generated answer for a particular video question. These video questions are specifically designed to target the temporal understanding in Video-LLMs. The dispreferred response is also model-generated answer for a particular video question. However, unlike the preferred response, we specifically make two types of perturbations to induce hallucinations in model responses as follows:
\begin{enumerate}
    \item We introduce special hallucinating elements in the prompt, making sure that it is still grounded to temporal sequence of events, forcing the model to provide responses that are not grounded to the original video.
    \item We produce perturbations in the frames itself, keeping the prompt same. These perturbations pertain to frames being corrupted with either lower resolution or distorted color sequences.
\end{enumerate}

Please refer to the supplementary materials (Section C) for details on hallucinating prompts and some qualitative examples on preference data generated using this method.

\subsection{Temporal Benchmark Development}
To provide a robust evaluation of temporal understanding grounded to the definition stated above, we construct a benchmark that goes beyond open-ended question-answer pairs by formulating Multiple Choice Question-Answer (MCQA) items. In this benchmark, the model must select the correct answer from multiple options, challenging its ability to discern fine-grained temporal dependencies within video sequences. Our specialized prompting method (a variation of the prompting method used to generate the preference data) guides the model in generating these multiple-choice question-answer pairs, emphasizing comprehension of event sequences and their interplay over the entire duration of the videos. An example of the benchmark is provided in supplementary section (D3).

\begin{table}[!htp]\centering
\caption{Benchmark Comparison. Only {\benchmark} \textbf{(Ours)} is the one that provides Shuffled and Reversed ordering of events on video for learning visual grounding and temporal dynamics}\label{tab:benchmark_stat}
\scriptsize
\scalebox{1.2}{
\begin{tabular}{lccc}\toprule
\textbf{} & \textbf{MCQs} & \textbf{Temporal} & \textbf{Shuffled} \\
\midrule
Next-QA & 8564 & \ding{55} & \ding{55} \\
Cinepile & 4941 & \ding{55} & \ding{55} \\
MSRVTT & 4134 & \ding{55} & \ding{55} \\
\midrule
TVBench & 2525 & \ding{51} & \ding{55} \\
VinoGround & 1000 & \ding{51} & \ding{55} \\
TempCompass & 7540 & \ding{51} & \ding{55} \\
VITATECS & 13.8k & \ding{51} & \ding{55} \\
Perception Test & 5000 & \ding{51} & \ding{55} \\
\hdashline
\addlinespace[2pt]
\orange\textbf{Ours} & \green\textbf{15k} & \green\ding{51} & \green\ding{51} \\
\bottomrule
\end{tabular}
}
\end{table}

Table \ref{tab:benchmark_stat} provides basic comparison between multiple benchmarks. Note that "shuffled" in the table refers to videos containing shuffled order of events rather than random shuffling of frames which may produce non-sensical and un-natural videos like in TVBench. We see that, only {\benchmark} is the one that contains 15k temporally targetted multiple-choice questions of this scale. To gain further insight into the effectiveness of {\benchmark} in comparison with different benchmarks, we conducted a t-SNE visualization analysis comparing the question embeddings of {\benchmark} with those of seven other benchmarks: CinePile, MSRVTT, NextQA, Perception Test, TempCompass, TVBench, and VinoGround (shown in Figure \ref{fig:t_sne}). Our analysis revealed a big gap in the current benchmarks (shown in the left part of Figure \ref{fig:t_sne}) and {\benchmark} bridges this gap (shown in the right part of Figure \ref{fig:t_sne}). Since MVBench \cite{li2024mvbench} is constructed with combination of these benchmarks, we decided skipping it in order to avoid redundancy.
\section{Experiments}
\label{sec:experiments}

\textbf{Benchmarks and Fine-Tuning Datasets}: 
To assess the improvement in the temporal capability of fine-tuned models, we perform extensive evaluations on multiple benchmarks such as Perception Test, TVBench, VinoGround, VITATECS, TempCompass, {\benchmark}, CinePile, MRVTT, and NextQA. Each of these benchmarks evaluates the model's capability in various aspects. For instance, the first six benchmarks focus on the temporal understanding of various aspects of a given video sequence. The next three focus on long-context, video description, and causal reasoning of the given video sequence. As a result, these evaluations comprehensively analyze the model’s proficiency across different video-based reasoning tasks.

We utilize FineVideo \cite{Farré2024FineVideo} as the preliminary dataset for constructing the preference dataset using {\finetune}, containing 15k QA-Pairs and 10k Videos (including both original and shuffled ones). In addition, to avoid overfitting, we incorporate preference data generated by LLaVA-Hound on ShareGPTVideo \cite{chen2025sharegpt4video}. We also use this dataset as one of the baselines (more details will be provided later). It is a 17k preference dataset designed to force the model to focus on the spatial content in the video sequence. Thus, for our case, we infused the preference dataset from LLaVA-Hound and that generated using {\finetune} to get a 32k preference dataset, which is then used for fine-tuning models. Moreover, we provide a baseline comparison for a combined, fused preference dataset incorporating the dataset generated from all the baseline methods as an additional ablation to our method. For this, we take the full 17k ShareGPTVideo preference dataset and a 7.5k randomly sampled subset of data generated by {\finetune} and TimeWarp-Implicit each.

\hspace{-12pt}\textbf{Baselines}: We perform extensive experiments on two baseline models: LLaVA-Hound and Video-LLaMA3. We utilize the former one because of its preference optimization infrastructure. Amongst all the recent SOTA models, we select Video-LLaMA3 because of its closeness to LLaVA-Hound in terms of processing the video sequence. For both these baseline models, we train them using five methodologies: Supervised Fine-Tuning (SFT), Base DPO method used in LLaVA-Hound to generate preference data from ShareGPTVideo (for simplicity we refer to this as Base-DPO or simply DPO), TimeWarp-Implicit, {\finetune} and Combination method.

\hspace{-12pt}\textbf{Implementation Details}: We align all the generated preference data according to the requirements of both the individual baseline model. Moreover, for a fair comparison, we consider models with 7B parameters. We fix the global batch size to 80 for all the baseline methodologies and models with a maximum learning rate of 5e-7, employing linear decay. Fine-tuning durations are set to approximately 650 for DPO (17k ShareGPTVideo preference data) and 1.2k for others (32k preference data). We uniformly sample 10 frames per video for our experiments. Models are trained on 5 A100 (80GB) GPU for all methods using bf16 precision and four gradient accumulation steps. The final training-related experiments will cost about 60 A100 GPU hours for LLaVA-Hound and about 25 A100 GPU hours for Video-LLaMA3. Lastly, to maintain computational efficiency, we use Low-Rank Adaptation (LoRA) for fine-tuning rather than full fine-tuning, with \textit{Rank = 128}, and \textit{Alpha = 256}. We leave the experiments on increasing the data and model size as future work.
\section{Results}
\label{sec:results}
\begin{table*}[hbt!]\centering
\caption{Performance comparison of baseline models, SOTA models, and methodologies against multiple benchmarks. SFT: Supervised Fine-Tuning dataset. DPO: Preference data constructed using ShareGPTVideo, Combined: Combination of randomly sampled data from above three methods, TimeWarp-Implicit: Preference data constructed using extended STIC method, TimeWarp-Explicit: Preference data constructed using our method. LLaVA and LLaMA indicate fine-tuning performed on LLaVA-Hound and Video-LLaMA3, respectively. \textbf{Bold} numbers indicate improvement from baseline. Green highlights represent the best performing method for that benchmark.}
\label{tab:results}
\scriptsize
\scalebox{0.9}{
\begin{tabular}{@{} >{\centering\arraybackslash}m{1pt} lcccccccccccccc @{}}
\toprule
 
&\multirow{2}{*}{\textbf{Methods}} &\multirow{2}{*}{\textbf{Perception Test}} &\multirow{2}{*}{\textbf{TVBench}} &\multicolumn{3}{c}{\textbf{VinoGround}} &\multicolumn{2}{c}{\textbf{TempCompass}} &\multicolumn{2}{c}{\textbf{TimeWar}} &\multirow{2}{*}{\textbf{CinePile}} &\multirow{2}{*}{\textbf{MSRVTT}} &\multirow{2}{*}{\textbf{Next-QA}}\\ \cmidrule(lr){5-7} \cmidrule(lr){8-9} \cmidrule(lr){10-11}
& & & &\textbf{Text} &\textbf{Video} &\textbf{Group} &\textbf{MCQs} &\textbf{Cap-Match} &\textbf{Normal} &\textbf{Shuffled} & & & \\
\midrule
\multirow{5}{*}{\rotatebox[origin=c]{90}{\textbf{LLaVA}}} &\textbf{SFT} & 52.61 &45.24 &24 &26 &6.2 &56.51 &60.1 &49.74 &48.96 &31.53 &67.6 &64.79 \\
&\textbf{DPO} &52.73 &44 &23.8 &26 &6.4 &56.96 &60.72 &49.92 &49.74 &32.91 &\green{83.35} &49.76 \\
\cdashline{2-15}
\addlinespace[2pt]
&\orange{\textbf{Combined}} \textit{(ours)} &\green{\textbf{57.58}} &42.03 &20.8 &25.6 &6 &\textbf{59.75} &\textbf{66.47} &\textbf{57.1} &\textbf{56.3} &\green{\textbf{33.11}} &\textbf{76.08} &\textbf{71.41} \\
&\orange{\textbf{TimeWarp-Implicit}} \textit{(ours)} &\textbf{56.69} &40.23 &\green{\textbf{33.2}} &21.2 &\green{\textbf{8}} &\textbf{55.7} &\green{\textbf{67.07}} &\green{\textbf{60.98}} &\green{\textbf{60.18}} &\textbf{33} &\textbf{73.17} &\green{\textbf{72.02}} \\
&\orange{\textbf{TimeWarp-Explicit}} \textit{(ours)} & \textbf{57.57} &\green{\textbf{47.39}} &22.2 &\green{\textbf{31.6}} &\textbf{6.6} &\green{\textbf{61.28}} &\textbf{60.95} &\textbf{55.36} &\textbf{55.14} &\textbf{32.02} &\textbf{78.86} &55.32 \\
\midrule
\midrule
\multirow{5}{*}{\rotatebox[origin=c]{90}{\textbf{LLaMA}}} &\textbf{SFT} &55.1 &43.57 &31.8 &29.2 &10.2 &63.92 &71.32 &53.9 &51.4 &34.04 &69.59 &69.81 \\
&\textbf{DPO} &56.44 &\green{43.61} &\green{34.8} &30.2 &11.8 &63.99 &72.85 &55.02 &{52.66} &34.89 &68.19 &69 \\
\cdashline{2-15}
\addlinespace[2pt]
&\orange{\textbf{Combined} \textit{(ours)}} &\green{\textbf{57.22}} &42.8 &\textbf{32} &\textbf{32.8} &\green{\textbf{12}} &\green{\textbf{65.76}} &\green{\textbf{73.32}} &\textbf{55.88} &\textbf{51.78} &\textbf{35.11} &\green{\textbf{71.14}} &\textbf{70.41} \\
&\orange{\textbf{TimeWarp-Implicit}} \textit{(ours)} &\textbf{56.68} &43.1 &\textbf{33.2} &\green{\textbf{35}} &\textbf{11.4} &\textbf{64.56} &\textbf{73.19} &\textbf{54.74} &51.4 &\textbf{34.89} &69.38 &68.2 \\
&\orange{\textbf{TimeWarp-Explicit}} \textit{(ours)} &\textbf{56.76} &43.53 &\textbf{33.4} &27.4 &7.8 &\textbf{65.57} &\textbf{72.92} &\green{\textbf{56.44}} &\green{\textbf{52.86}} &\green{\textbf{35.66}} &\textbf{70.85} &\green{\textbf{71.61}} \\
\midrule
\midrule
\multirow{4}{*}{\rotatebox[origin=c]{90}{\textbf{SOTA}}} & \textbf{Qwen2.5-VL} &64.13 &44.13 &36.2 &27.6 &11.4 &\textbf{71.07} &75.04 &61.12 &55.56 &\underline{39.34} &38.6 &\underline{76.22} \\
& \textbf{InternVideo} & 67.49 & \textbf{50.77} &39.8 & 28.59 & 11.6 &69.41 & \textbf{79.7} & \textbf{66.76} & \textbf{60.44} & \textbf{40.29} & 41.34 & \textbf{79.4} \\
& \textbf{VideoChat-Flash} &\textbf{69.14} &{50.59} &\textbf{55.8} &\textbf{33.4} &\textbf{21.8} & 71.01 & 77.11 & 66.1& 57.24 &32.54 &39.35 &70.38 \\
& \textbf{LLaVA-OneVision} &56.53 &43.57 &39.4 &28.59 &{15} &64.17 &71.98 &55.06 &49.59 &35.13 &43.94 &76.07 \\
\bottomrule
\end{tabular}
}
\end{table*}

\textbf{Overview}: Our primary results are displayed in Table \ref{tab:results}, where we divide it into three sections, two of which are for the two base models viz LLaVA-Hound (represented as LLaVA) and Video-LLaMA3 (represented as LLaMA). The third section provides performance overviews of current SOTA models. For the first two sections, each section further consists of five methods used to generate the dataset and fine-tune on downstream tasks. Our results show a consistent and significant improvement in the performance of baseline models across almost all seven benchmarks.

\hspace{-12pt}\textbf{Finetuning Performance}: We observe almost 7.5\% absolute improvement in temporal tasks and average performance improvement of about 4.5\% across all temporal benchmarks for LLaVA-Hound baseline. We also observe a similar trend with Video-LLaMA3 with average performance an absolute improvement of 2.5\% across all temporal benchmarks. Beyond temporal tasks, fine-tuning with \textbf{\finetune} also leads to notable improvements in longer video benchmarks, such as CinePile, where we observe a performance increase of around 2.5\% and 1.62\% for both the base models. This enhancement suggests that even with a reduced frame rate (10 FPV), our dataset equips LLaVA-Hound and Video-LLaMA3 with the necessary temporal granularity to handle long-video understanding effectively. The result highlights the broader applicability of preference data generated from {\finetune}, indicating that our dataset can address limitations related to temporal dynamics in various video lengths and types.

\hspace{-12pt}\textbf{Outliers and Anomalies}: However, certain limitations remain, as seen in the performance gap for the NextQA benchmark (a causal and counterfactual reasoning benchmark). Despite achieving a nearly 6\% improvement relative to the DPO baseline for LLaVA-Hound, the fine-tuned model still falls short of the results achieved by training on SFT data. This indicates that focusing on the broader perspective on temporal \textit{understanding} may lead to a decrement in temporal \textit{reasoning} of models, which requires more fine-grained analysis of the events. However, the performance of Video-LLaMA3 improved by 1.8\%, indicating newer models are robust and retain the capability for improvement using preference optimization without compromising the \textit{reasoning} capabilities.

\hspace{-12pt}\textbf{SOTA Comparison}: In addition to the baseline models, Table \ref{tab:results} presents the performance of more recent and advanced models, including VideoChat-Flash \cite{li2024videochatflash}, InternVideo \cite{wang2025internvideo}, Qwen2.5-VL \cite{qwen2.5-VL}, and LLaVA-OneVision \cite{onevision} to provide a comprehensive comparison. We observe that these newer models achieve higher performance compared to our fine-tuned models. However, we attribute these performance gains to architectural enhancements and specific training strategies employed by these models. Moreover, the performance of the weaker models is observed to be increasingly comparable to that of the state-of-the-art (SOTA) models across all benchmarks after fine-tuning on the preference dataset generated using our method. This suggests that targeted fine-tuning alone allows these models to reduce the performance gap significantly with respect to SOTA.

\hspace{-12pt}\textbf{{\benchmark}}: Interestingly, we observe a substantial performance drop, averaging approximately 4\%, when comparing Normal and Shuffled Videos within our proposed benchmark ({\benchmark}). This provides empirical evidence indicating a lack of temporal \textit{understanding} in current Video-LLMs. Furthermore, we observe that Video-LLaMA3 exhibits an average performance drop of approximately 3\%, whereas LLaVA-Hound shows only a minor decline of about 0.5\%. This suggests that models fine-tuned using our proposed method demonstrate greater robustness, exhibiting a smaller performance degradation on tasks requiring temporal understanding. Finally, even when comparing LLaVA-Hound and Video-LLaMA3 directly, we observe a notable difference in the average performance drop. We attribute this to the fact that more recent models benefit from improved training strategies, including enhanced video-text alignment. Consequently, these models possess more fine-grained temporal reasoning capabilities, making the performance drop more pronounced when temporal information is disrupted.
\section{Ablations}
This section provides an in-depth analysis of the ablation studies conducted to validate the effectiveness of the proposed method and design choices. We systematically evaluate the impact of various components of the TimeWarp methodology on the performance of the models. This includes the use of preference optimization technique other than DPO. Moreover, We aim to augment our previous findings with an extensive probing of the model's understanding of temporal nuances before and after finetuning using the TimeWarp temporal preference dataset. This section also includes some fine-grained probes aiming to test the model's reasoning of fine-grained temporal concepts beyond existing benchmarks. 

\subsection{Kahneman-Tversky Optimization (KTO)}
The KTO framework proposed by \cite{ethayarajh2024kto} advances two central hypotheses regarding its comparative effectiveness with established preference optimization methods:
\begin{itemize}
    \item \textbf{SFT+KTO vs. SFT+DPO:} Through empirical studies on LLaMA-7B, 13B, and 30B, the authors demonstrate that models trained with KTO following supervised fine-tuning achieve performance levels comparable to those optimized with DPO, despite KTO leveraging a comparatively weaker training signal. To probe the validity of this observation in a multimodal setting, we conduct an initial study on LLaVA-Hound, adapting the existing DPO preference dataset into a KTO-compatible training signal.
    \item \textbf{Pretrained+KTO vs. SFT+KTO.} The second claim is that applying KTO directly on pretrained models, bypassing supervised instruction-tuning, yields results that remain competitive with the SFT+KTO pipeline. This finding, if substantiated more broadly, could challenge the assumption that instruction-tuning is a prerequisite for preference-based optimization. We defer the empirical verification of this hypothesis in the multimodal temporal reasoning setting to future work.
\end{itemize}

To the best of our knowledge, a comparative study of KTO vs DPO has not been reported in multimodal settings. We therefore aim to provide some initial findings along this line.

We repurpose the existing preference data into a KTO-compatible format by decomposing each video–question pair into two samples labeled $True$ and $False$. Using the shuffled videos from the TimeWarp-Explicit generation process, we construct additional samples by reversing the original preferences (i.e., negative for the original becomes positive for the shuffled). This procedure yields four KTO samples for every DPO sample (two from the original and two from the shuffled videos). From this pool, we randomly select 15k KTO samples to fine-tune LLaVA-Hound. To ensure comparability with the DPO experiments, we adopt the same training configuration.

\begin{table}[hbt!]
\centering
\caption{Comparison of DPO vs KTO for LLaVA-Hound across VinoGround and TempCompass benchmarks}
\label{tab:kto_dpo_comparison}
\vspace{0.3cm}
\begin{tabular}{lcccc}
\toprule
\multirow{2}{*}{\textbf{Method}} & \multicolumn{2}{c}{\textbf{VinoGround}} & \multicolumn{2}{c}{\textbf{TempCompass}} \\
\cmidrule(lr){2-3} \cmidrule(lr){4-5}
& \textbf{Text} & \textbf{Video} & \textbf{Multi-Choice} & \textbf{Cap-Match} \\
\midrule
\textbf{SFT} & \green{24.1} & 26.5 & 56.52 & 60.1 \\
\textbf{DPO} & 23.7 & 26.02 & \green{61.28} & \green{60.95}\\
\textbf{KTO} & 24.05 & \green{26.55} & 53.48 & 55.88 \\
\bottomrule
\end{tabular}
\end{table}

The comparative results for KTO vs DPO on LLaVA-Hound are shown in Table \ref{tab:kto_dpo_comparison}. Surprisingly, LLaVA-Hound finetuned on KTO temporal preference data even underperforms with respect to LLaVA-Hound, contrary to results reported in the KTO Paper \cite{ethayarajh2024kto}. Moreover, KTO suffers from unstable training, with loss functions not converging properly, indicating underfitting even after a sufficient number of training steps (1.4k).

We provide a detailed analysis with qualitative examples explaining this contradiction in Appendix F. Our preliminary findings suggest that the temporal preferences in our dataset may be too subtle for KTO to effectively leverage, resulting in the observed underfitting. Further verification on additional models such as Video-LLaMA3 remains as future work.

\subsection{Binary Temporal Ordering: Near vs Far?}
In this ablation, we aim to probe the fine-grained temporal understanding of video events in a binary setup, different from our original Multi-choice based evaluations. While proficiency in multi-choice evaluations provides sufficient evidence of the model's overall temporal understanding, it would be worthwhile investigating whether the model can reproduce a similar performance in a stepwise exploratory setup.

We sample a set of 500 videos from the FineVideo dataset \cite{Farré2024FineVideo} along with their associated event captions and corresponding timestamps. The chosen videos are constrained to have at least 4 event captions. We fix a middle event (the middle caption in the caption sequence) and sample from the remaining captions across 3 categories:

\begin{itemize}
    \item \textbf{Near:} The middle event caption paired with one of its adjacent (1-hop) event captions. (Event time separation $\sim$ 10-20s)
    \item \textbf{Moderately Far:} The middle event caption paired with one of its 2-hop separated event captions. (Event time separation $\sim$ 20-30s)
    \item \textbf{Very Far:} The middle event caption paired with any one of its $\ge$ 2-hop separated event captions. (Event time separation $\ge$ 30s)
\end{itemize}

For each temporal order category, we define two subtypes (Before and After) resulting in six subcategories in total. Given a set of event captions $E$, for each pair of distinct events $e_A, e_B \in E$ with the correct order $o(e_A, e_B)$ (where $o : E \times E \rightarrow {Before, After}$), we \textbf{generate} four binary statements:

\begin{itemize}
\item $e_A [o(e_A, e_B)] e_B$: $e_A$ occurs before $e_B$ (True)
\item $e_A [\sim o(e_A, e_B)] e_B$ (False): $e_A$ occurs after $e_B$
\item $e_B [\sim o(e_A, e_B)] e_A$ (True): $e_B$ occurs before $e_A$
\item $e_B [o(e_A, e_B)] e_A$ (False): $e_B$ occurs after $e_A$
\end{itemize}

Each statement is posed to the model, which must respond with Yes/No. Thus, each temporal order category has eight binary statements (four for Before, four for After).

To evaluate model correctness on a given video and subcategory, we experiment with different criteria. Both LLaVA-Hound and Video-LLaMA3 perform nearly perfectly when only 1–2 statements need to be correct, largely by defaulting to “No” on False statements. However, when all four statements must be correct, accuracy drops below 1\%, highlighting the difficulty of capturing temporal order consistently. For balanced evaluation, we require at least 3 out of 4 statements correct at the subcategory level, and at least 5 out of 8 at the category level.

\begin{figure}[hbt!]
\centering
\includegraphics[width=0.9\linewidth]{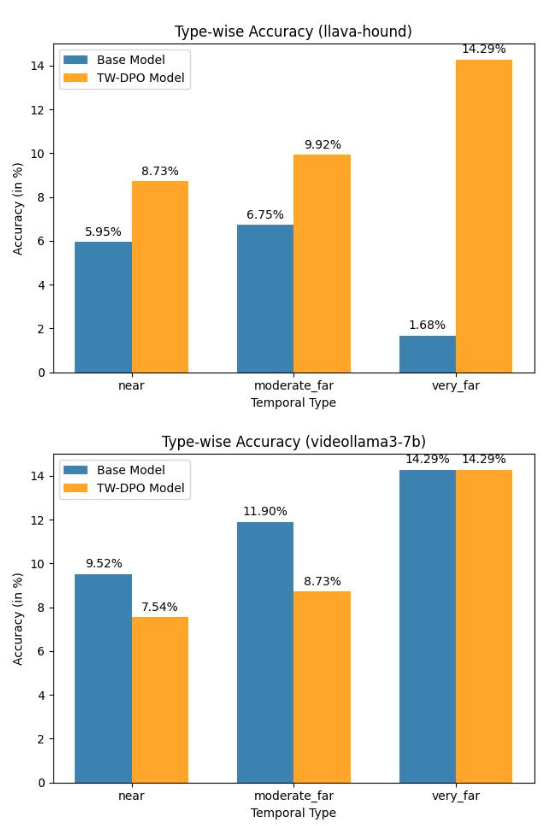}
\caption{Binary temporal order evaluation results for temporal order categories.}
\label{fig:temporal_order_categories}
\end{figure}

\begin{figure}[hbt!]
\centering
\includegraphics[width=0.9\linewidth]{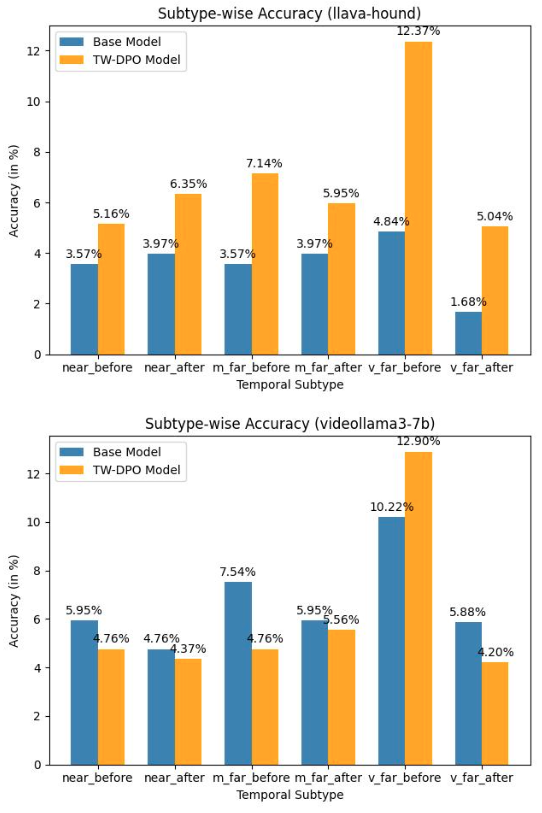}
\caption{Binary temporal order evaluation results for temporal order subcategories.}
\label{fig:temporal_order_subcategories}
\end{figure}

Figures \ref{fig:temporal_order_categories} and \ref{fig:temporal_order_subcategories} present results for both models. LLaVA-Hound shows a consistent 2-3\% improvement across categories and subcategories, suggesting that preference fine-tuning benefits models without strong temporal priors. In contrast, Video-LLaMA3 shows slight decreases in most categories. This may indicate that the LLM backbone already leverages commonsense temporal priors, especially for moderate ranges ($\leq$30s), and that preference training disrupts these priors by forcing reliance on video evidence. The improvement in the very far category supports this view. Finally, we observe a notable asymmetry: performance improves in the very far before subcategory but not in very far after. This pattern is present in both base models and becomes more pronounced after fine-tuning.
\section{Conclusion}
\label{sec:conclusions}

This work tackled a critical gap in Video-LLMs: the underdeveloped capability to understand temporal dynamics within video sequences. In an attempt to address this, we developed a comprehensive, large-scale synthetic data generation pipeline, {\finetune}, specifically tailored to enhance the model’s temporal comprehension by grounding responses in both visual and temporal intricacies. Our method is meticulously designed to support models in learning the interplay between events and scenes, enabling deeper insights into sequences and transitions over time. Moreover, we provide a method to indirectly generate temporal preference data for Video modality.

Our comprehensive experiments across a suite of benchmarks demonstrate that {\finetune} significantly outperforms existing methodologies such as Base-DPO, achieving notable gains across almost all the temporal benchmarks. This performance boost underscores the potential of our approach to advance the temporal understanding capabilities of Video-LLMs, marking a substantial step forward in the pursuit of robust spatiotemporal intelligence.

\section{Acknowledgments}
The project is partially supported by US NSF RI grants \#1750082 and \#2132724. 
We thank the Research Computing (RC) at Arizona State University (ASU) for their generous support in providing computing resources \cite{jennewein2023sol}.
The views and opinions of the authors expressed herein do not necessarily state or reflect those of the funding agencies and employers. YY holds concurrent appointments at Arizona State University and as an Amazon Scholar. This paper describes work performed at Arizona State University and is not associated with Amazon.

{
    \small
    \bibliographystyle{ieeenat_fullname}
    \bibliography{main}
}

\onecolumn
\appendix

\begin{center}
{\Large \textbf{Appendix}}
\end{center}

\section{Overview}
This section will provide more details regarding various aspects of the main paper. We begin by providing and discussing briefly the background information for Direct Preference Optimization and alignment fine-tuning. Following that, we provide qualitative examples and more details about data generated using TimeWarp-Implicit. We then provide the task instructions used for generating data using \textbf{TimeWarp} and conclude with some limitations and future work.

\section{Preliminaries}
To improve LLM alignment with human preferences, RL-based fine-tuning \cite{christiano2017deep} is applied, typically, after supervised fine-tuning (SFT) to fine-tune on the downstream tasks. This process begins with an initial reward function $r(x,y)$ for a given input sequence $(x, y)$ which is then trained according the the preference data. The more preferred response $y$ is expected to result in higher reward $r(x, y)$. As a result, the corresponding objective is to maximize Eq. \ref{eq:rl_loss},
\begin{equation} 
    \centering
    \label{eq:rl_loss}
    L(\theta) = \mathbb{E}_{x \sim D, y \sim p_{\theta}(.|x)}[r(x,y)] -\lambda \hspace{3pt} \mathbb{E}_{x \sim D}KL(p_{\theta}(.|x) || p_{ref}(.|x))
\end{equation}
where $x \sim D$ is sampled from a given distribution $D$ and the KL regularization term prevents the new model $p_{\theta}$ from deviating too much from the reference model $p_{ref}$, with $\lambda > 0$ as the regularization parameter.

Since training the reward function is challenging in practice, Direct Preference Optimization \cite{rafailov2024direct} simplifies this process using predefined preference dataset $S_{ref} = \{x^{(i)}, y_{w}^{(i)}, y_{l}^{(i)}\}_{i \in N}$, where $y_{w}^{(i)}$ denotes the preferred response and $y_{l}^{(i)}$ denotes the dispreferred response given the same prompt $x^{(i)}$. As a result, the corresponding objective is to maximize Eq. \ref{eq:dpo_loss},
\begin{equation}
    \centering
    \label{eq:dpo_loss}
    L_{DPO}(\theta, \theta_{ref}) = \mathbb{E}_{(x, y_{w}, y_{l}) \sim S_{pref}}\left[l\left(\lambda \hspace{3pt} log\left(\frac{p_{\theta}(y_{w}|x)}{p_{\theta_{ref}}(y_{w}|x)}\right) - \lambda \hspace{3pt} log\left(\frac{p_{\theta}(y_{l}|x)}{p_{\theta_{ref}}(y_{l}|x)}\right)\right)\right]
\end{equation}
where $l(t) = log(1 + \exp(-t))$ is the logistic loss function and $\theta_{ref}$ is the reference model.

\section{TimeWarp-Implicit}
In this section, we provide some qualitative examples for our proposed TimeWarp-Implicit for generating preference dataset. As mentioned in the paper, we use two types of perturbations to generate preferred and dispreferred responses: Prompt-Based and Frame-Based. Following are a subset of "bad" prompts that induce spatiotemporal hallucinations in the response of models as a means of generating dispreferred responses. To minimize redundancy, we only show one example of the preference data generated by Video-LLaMA3 only. We have obtained similar results for LLaVA-Hound as well.

\begin{itemize}
    \item Describe the video with imaginative sequences of events that may unfold over time.
    \item Enrich the video narrative by adding hypothetical events or interactions that could occur between characters or objects.
    \item Suggest and detail practical sequences or interactions that could logically happen within the video's timeline.
    \item Incorporate elements that, though absent, would seamlessly fit into the temporal flow of the video.
    \item Imagine and describe additional everyday activities or interactions taking place just out of frame in the video.
    \item Augment the video with details of potential events or interactions that are plausible over time.
    \item Conceive of and detail natural phenomena, such as weather changes or animal movements, that could realistically occur during the video's duration. Make the description affirmative.
\end{itemize}




\section{Task Instructions for TimeWarp}
\subsection{Open-Ended QA-Pair Generation}
\newpage
\begin{figure}[hbt!]
    \centering
    \includegraphics[width=\linewidth]{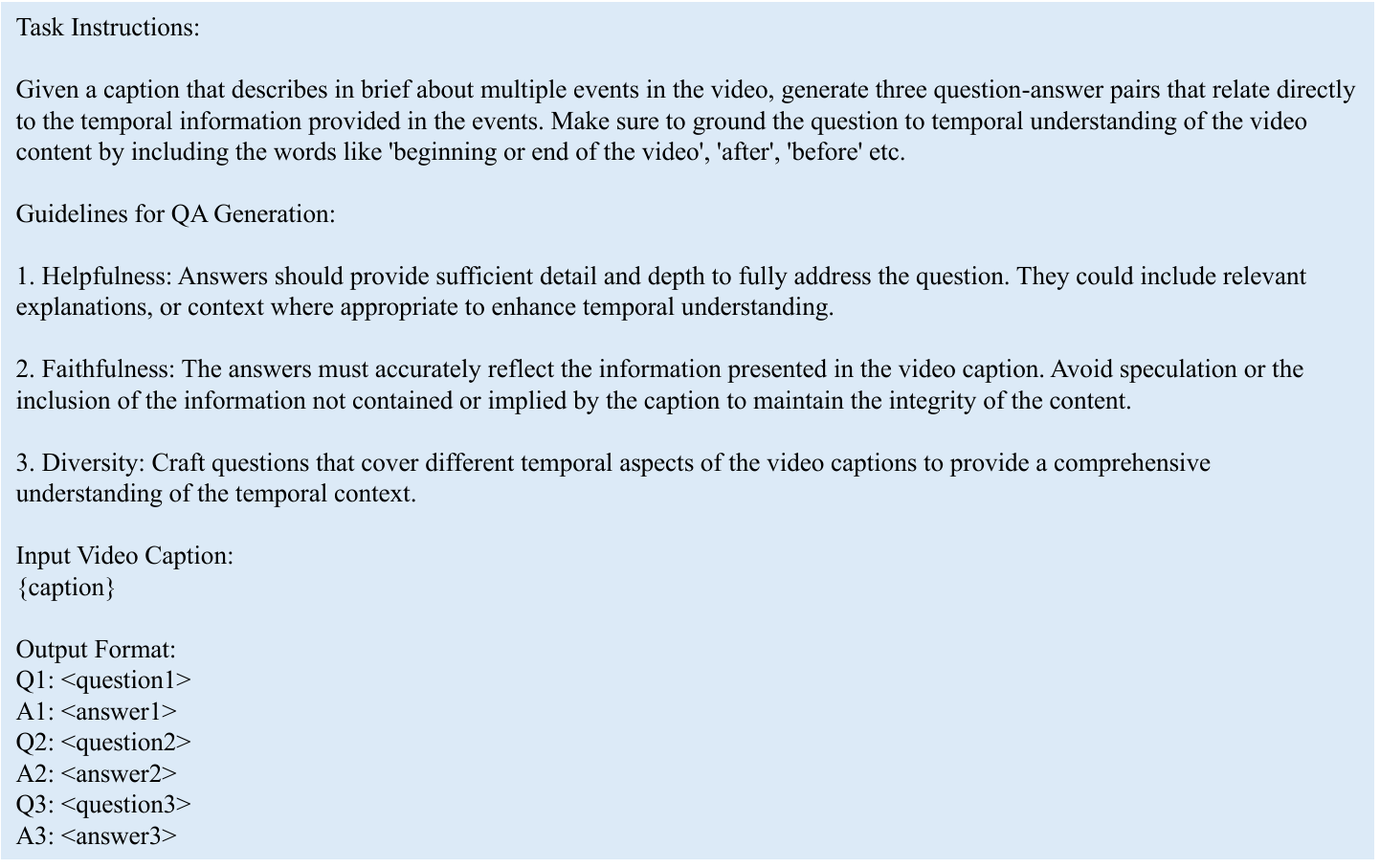}
    \caption{Instruction given to GPT-4o-Mini for generating the open-ended Question-Answer pairs}
    \label{fig:pref_task_instructions}
\end{figure}

\newpage
\subsection{Multiple Choice QA-Pair Generation}
\begin{figure}[hbt!]
    \centering
    \includegraphics[width=\linewidth]{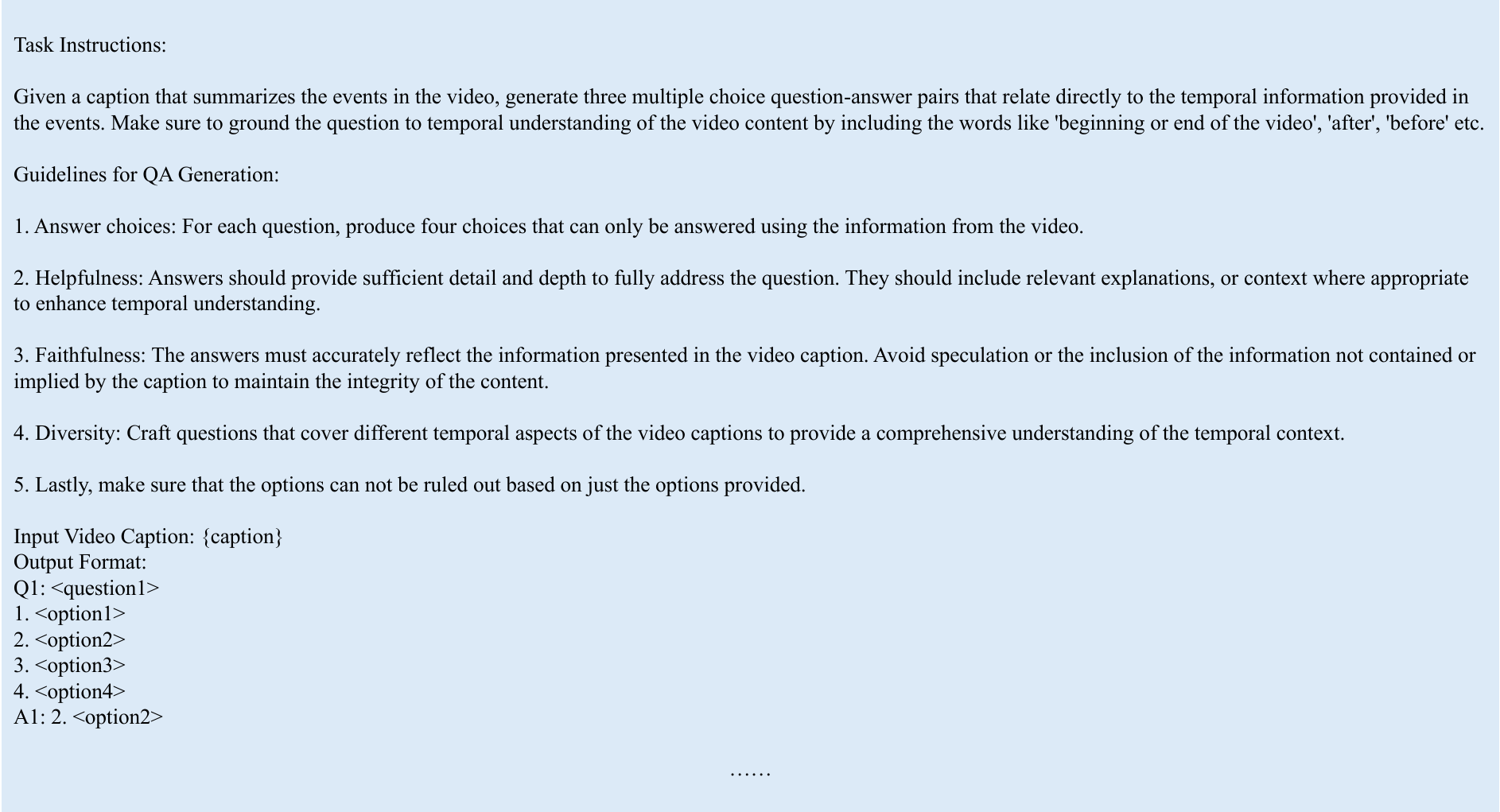}
    \caption{Instruction given to GPT-4o-Mini for generating multiple choice Question-Answer pairs}
    \label{fig:pref_task_instructions}
\end{figure}


\section{Limitations and Future Work}
While data generated using \textbf{TimeWarp} demonstrates substantial improvements across a variety of benchmarks, some limitations persist that indicate directions for future enhancement. First limitation stems from the use of basic clip captions from FineVideo, which may not fully capture the intricate dynamics between events in a video sequence. By augmenting these captions to include more detailed event transitions and contextual cues, future iterations of \textbf{TimeWarp} could provide richer temporal context, thereby improving the model's ability to discern subtle changes and relationships between scenes. Second, applying \textbf{TimeWarp} to other advanced architectures and SOTA, such as Video-Chat and InternVideo, will reveal further insights into how different model architectures can process and retain temporal information. Such adaptations could also open up opportunities for multi-modal datasets, potentially combining audio and text cues with video to provide a more comprehensive framework for real-world video understanding. However, our method can be extended easily to any other model architecture that can use both SFT and preference fine-tuning.

\section{KTO Underfitting Analysis}
\begin{figure}[hbt!]
    \centering
    \includegraphics[width=\linewidth]{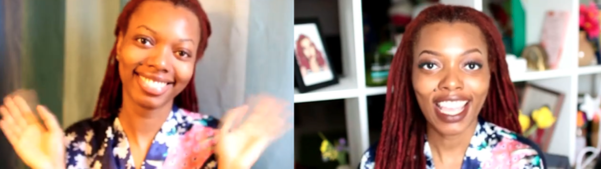}
    \caption{A sample video sequence from the KTO preference dataset showing successive events. In the first frame, the woman smiles and waves her hand. In the second frame, she discusses her skincare routine.}
    \label{fig:kto_example}
\end{figure}

To investigate why KTO underperforms compared to DPO on our temporal preference data, we examine a qualitative example from the KTO dataset. Figure \ref{fig:kto_example} shows a pair of positive and negative preferences from the same video-question pair.

The corresponding positive and negative preferences are:
\begin{itemize}
    \item \textbf{Prompt:} What does the person do at the beginning of the video before discussing her skincare routine?
    \item \textbf{Positive:} At the beginning of the video, the person is seen smiling and waving to the camera, which sets a friendly and engaging tone before she delves into her skincare routine.
    \item \textbf{Negative:} At the beginning of the video, the person is smiling and looking down before she starts discussing her skincare routine.
\end{itemize}

The positive and negative samples are distinct (waving vs. looking down), yet remain semantically similar. According to \cite{ethayarajh2024kto}, KTO updates leave $\pi_\theta$ unchanged when an example is too difficult to learn from. While this property helps avoid noisy updates, it also risks discarding challenging but informative samples, potentially leading to underfitting.

In our case: (i) positive and negative preferences are temporally distinct but visually similar, and (ii) the KTO dataset additionally includes shuffled video-prompt-response tuples. As a result, a substantial fraction of training samples may be too difficult for the model to differentiate, increasing the likelihood that KTO ignores them during training, causing the observed underfitting.

\end{document}